\def\year{2021}\relax
\documentclass[letterpaper]{article} 
\usepackage{aaai21}  
\usepackage{times}  
\usepackage{helvet} 
\usepackage{courier}  
\usepackage[hyphens]{url}  
\usepackage{graphicx} 
\def\UrlFont{\rm}  
\usepackage{natbib}  
\usepackage{caption} 
\nocopyright
\usepackage{amsmath}
\usepackage{diagbox}
\usepackage{multirow}
\usepackage[switch]{lineno}
\usepackage{amsfonts,amssymb}

\title{Deep Template Matching for Pedestrian Attribute Recognition \\with the Auxiliary Supervision of Attribute-wise Keypoints}
\author{
	Jiajun Zhang*,\textsuperscript{\rm 1}
	Pengyuan Ren*,\textsuperscript{\rm 1}
	Jianmin Li\textsuperscript{\rm 2}\\	
}
\affiliations{
	\textsuperscript{\rm 1}SensingTech Ltd., Beijing\\
	\textsuperscript{\rm 2}Tsinghua University\\
	\{zhangjiajun,renpengyuan\}@sensingtech.com.cn, lijianmin@mail.tsinghua.edu.cn
}
\begin{document}
\maketitle
\begin{abstract}
		Pedestrian Attribute Recognition (PAR) has aroused extensive attention due to its important role in video surveillance scenarios. In most cases, the existence of a particular attribute is strongly related to a partial region. Recent works design complicated modules, e.g., attention mechanism and proposal of body parts to localize the attribute corresponding region. These works further prove that localization of attribute specific regions precisely will help in improving performance. However, these part-information-based methods are still not accurate as well as increasing model complexity which makes it hard to deploy on realistic applications. In this paper, we propose a Deep Template Matching based method to capture body parts features with less computation. Further, we also proposed an auxiliary supervision method that use human pose keypoints to guide the learning toward discriminative local cues. Extensive experiments show that the proposed method outperforms and has lower computational complexity, compared with the state-of-the-art approaches on large-scale pedestrian attribute datasets, including PETA, PA-100K, RAP, and RAPv2\textsubscript{$zs$}.
\end{abstract}

\begin{figure}[ht]
	\begin{center}
		\includegraphics[scale=0.68]{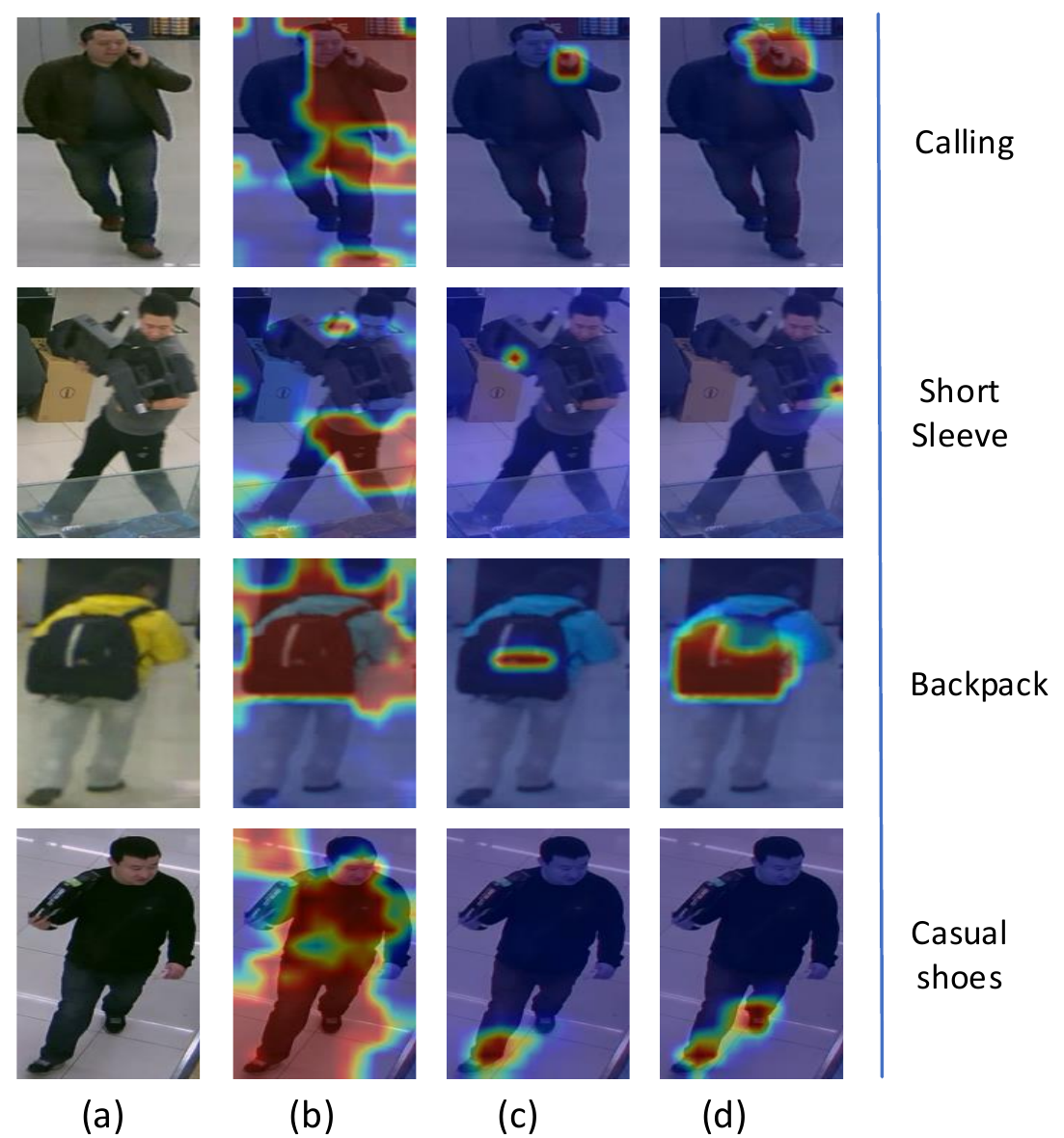}
	\end{center}
	\caption{Specific attribute attentive area. (a) Example images of pedestrian attribute recognition from RAP dataset \cite{RAPv1}. (b-d) Visualization of DTM output heatmaps, followed by global average pooling, global max pooling and global max pooling with attribute-wise keypoints supervision. Visualized attributes are calling, short sleeve, backpack, and casual shoes.}
	\label{fig1}
\end{figure}

\section{Introduction}
Pedestrian Attribute Recognition (PAR) has been widely used in many video surveillance applications, e.g., person re-identification \cite{reid1,reid2}, person retrieval \cite{retrieval1} and etc. Given a person's image, PAR aims to predict a group of attributes, e.g., gender, age, clothing, color, and attachment to describe the characteristic of this person, hence this task can be treated as a multi-label classification problem. As the prosperous development of deep learning, various methods based on convolutional neural networks (CNNs) have achieved great success. However, this task is still challenging due to factors such as occlusion, low resolution, illumination, multi-view, and unbalanced data distribution. 

Intuitively, different kinds of attributes have strong spatial relationships with different human body parts. During the process of attribute recognition, most attributes are only related to local regions, rather than global. For example, clothing can be determined by upper or lower body parts, hat and hair are around head, attachments are around hands. The point in improving attribute recognition performance is to capture the local characteristics which reflect the difference between attributes and eliminate the influence of background and interferer. Recent works append region attentive modules to deep neural network, e.g., body part proposal, segmentation, attention, STN. For example, \cite{PGDM,GRL} introduce external pose guided branch which locate body part regions, \cite{PA100K,AttentionAAAI20} apply attention mechanism to enhance the attribute-related region. The above research proved the importance and effectiveness of using attribute specific region information in attribute recognition. However, modules proposed by recent research bring in extra model complexity and computation cost, which make the model difficult to deploy in real-time applications. To keep model inference at high speed, we hope to use a less computationally intensive method to capture partial information of human body parts, while maintaining the performance of model at a high level.

In this paper, we introduce a training method called Deep Template Matching (DTM) with the auxiliary supervision of attribute-wise keypoints (AWK). Template matching is a classical method in the field of image recognition. Templates can extract the characteristics of local region and eliminate the irrelevant background by using the local output of template matching, instead of complicated region proposal modules. In the field of deep learning, convolution combined with pooling layer can achieve the effect of template matching, we call it DTM for short. Training DTM is still challenging due to the reasons: (1) Initialization with random weights may result in the randomness of local extrema which may leads model parameters update in the wrong location and increase the difficulty of training templates. As shown in Fig. \ref{fig1}(c) short sleeve, templates extract features on the person's attachment, rather than elbow. In Fig. \ref{fig1}(b) the inaccurate positioning of attribute related area is more obvious. (2) Naive template matching identify the matches using local maxima, however, limited matching places may not enough to discriminate hard samples, further, results in model overfitting and poor generalization. As shown in Fig. \ref{fig1}(c) casual shoes and backpack, the feature extracted is limited and the template only learned the characteristics of one foot. To solve these problems, we propose an auxiliary training method with human pose keypoints supervision for local features. AWK supervision improves the performance of DTM significantly and does not introduce model complexity at the inference stage. As illustrated in Fig. \ref{fig1}(d), visualization of attribute specific area shows that DTM extract features in a more accurate and reasonable region than others. Compared with existing approaches, our works achieve the state-of-art performance on most datasets, meanwhile, has less model complexity and computation cost. 

The contributions of this paper are as follows: 
\begin{itemize}
	\item We propose to use Deep Template Matching (DTM) as classifier to extract attribute features. To the best of our knowledge, there are no relevant works to implement PAR with template matching. 
	\item We propose to leverage human pose keypoints as auxiliary information to supervise DTM learning for proper attribute specific region.
	\item Experiments on major benchmarks, i.e. RAP \cite{RAPv1}, PA-100K \cite{PA100K}, PETA \cite{PETA} and RAPV2\textsubscript{$zs$} \cite{Rethinkingzeroshot} shows that our method have achieved state-of-the-art performance without introducing extra model complexity and computation cost.
\end{itemize}

\section{Related Work}
\par\textbf{Pedestrian Attribute Recognition} \hspace{3pt}Early pedestrian attributes recognition methods focus on hand-crafted features and attributes relations, such as HOG \cite{HOG}, SVM \cite{svm,svmreid}. However, these traditional methods are far from satisfactory in realistic applications. In recent years, convolution neural network has achieved great success in many computer vision tasks, including pedestrian attribute recognition. Wang et al. \cite{PARsurvey} give a detailed review of existing works. Li et al. \cite{deepmar} treated PAR as a multi-label classification problem and proposed weighted cross-entropy loss to handle the unbalance among attributes. The performance of global image-based methods is limited due to the lack of consideration for the fine-grained feature. Li et al. \cite{PGDM} utilize a pose estimation model to get human part regions. Liu et al. \cite{LGNet} use EdgeBoxes \cite{edgebox} to generate regions proposal to locate the region of interest.  Although these methods improve the overall performance significantly by using the body part information from an external part localization module, they also bring model complexity and time-consuming in both train and inference processes. Liu et al. \cite{PA100K} introduced attention modules for multi-scale features. Sarfraz et al. \cite{vespa} induced view-specific information into the attribute prediction units. However, visual attention mechanism based methods are limited, since the weights or masks generated may be confused by surroundings and background. Wang et al. \cite{JRL} proposed a CNN-RNN based method capable of jointly learning image-level context and attribute level sequential correlation. Zhao et al. \cite{GRL} recognize human attributes by group step by step to pay attention to both intra-group and inter-group relations. These CNN-RNN based methods are time-consuming and hard to use in real-time applications. \cite{gcn} use two Graph Convolutional Networks (GCN) to explore the correlations among attributes and regions. In addition to localizing attribute specific region, finding the correlation between attributes is also an aspect that many research concerns. From previous studies, there is still a long way to go of improving PAR performance in many aspects. At the same time, the complex design of modules leads to a timeliness issue which is inevitable in realistic deployment.

\textbf{Human Pose Estimation} \hspace{3pt}Human Pose Estimation (HPE) aims to estimate human joints (also known as keypoints, e.g., elbows, wrists, hip, knees, etc.) in images or videos. Recent research adopted ConvNets as the main building block has yielded drastic improvements on benchmarks \cite{deeppose,openpose,hourglass,CenterNet}.
Pose information provides us prior knowledge of the interest region and has been utilized in many existing PAR works. Zhang et al. \cite{Two_stream} add a landmark detection branch to jointly learn the attributes and landmarks. Unlike the fashion dataset \cite{deepfashion}, PAR datasets are not annotated with landmark information, we have to utilize a well-trained HPE model to generate keypoints as prior knowledge. Different from multi-person pose estimation, which needs to handle both detection and localization tasks since there is no prompt of how many persons in the input images, HPE can be treated as 2D single person pose estimation for images in pedestrian attribute datasets, which is easier than the multi-person case and most state-of-art HPE works can get satisfactory results.

\begin{figure*}[ht]
	\begin{center}
		\scalebox{1}[0.90]{
		\includegraphics[scale=0.3]{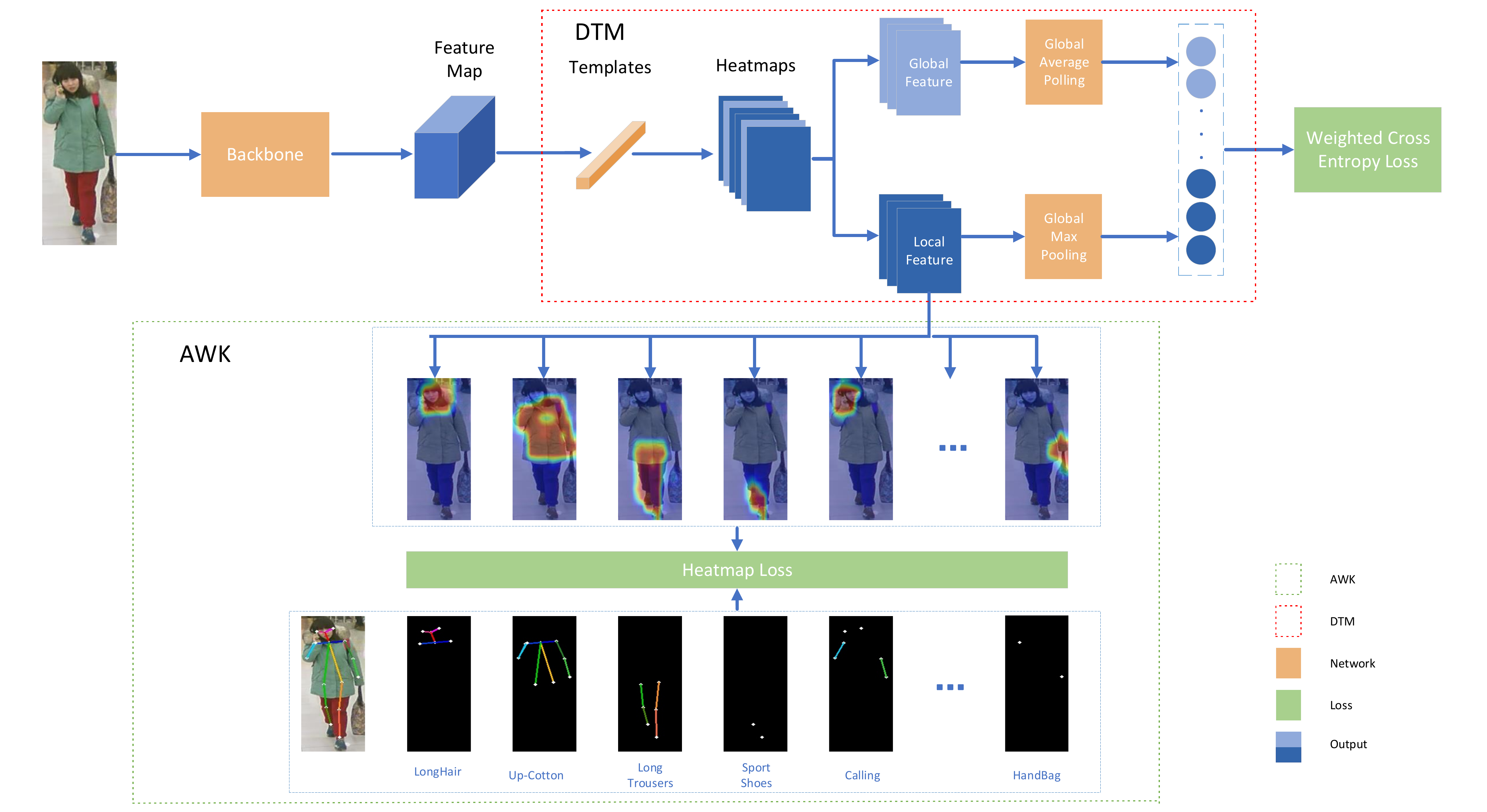}}
	\end{center}
	\caption{The overview framework consists of a Main Net and an AWK supervision training block. The input image is fed into the Main Net and the output of Deep Template Matching is divided into two sets, one for global feature (e.g., age, shape, gender), one for local feature (e.g., cloth, attachment, accessory), then concatenate the features after pooling layers. Local feature heatmaps are fed into the auxiliary supervision training block, which calculates heatmap loss for the guidance of template localization. Note that, the auxiliary block is applied to the training process only.}
	\label{fig3}
\end{figure*}

\section{Proposed Method}
The network we proposed consists of two parts while training, as illustrated in Fig. \ref{fig3}. One is the Main Network, which is a plain CNN architecture. For the Main Network, we treat the pedestrian attribute recognition as a multi-label classification task, and Deep Template Matching (DTM) is utilized to extract features. One is the auxiliary supervision training block, which is applied in training process only. For auxiliary supervision block, attribute-wise keypoints (AWK) are applied to output heatmaps as an auxiliary supervision tool for guiding template localization. Specifically, given a PAR dataset $\mathbb{D} = \{(I_i,y_{ij})\}$, where $y_{ij}\in\{0,1\}, i=\{1, 2,..., N\}, j=\{1, 2,..., J\}$, $y_{ij}$ indicates the ground truth of $j$-th attribute of $i$-th image, $N, J$ denotes the number of images and attributes respectively, PAR aims to predict attributes $\hat{y}_{ij} \in \{0,1\}$. The whole network is end-to-end trained using a loss function which is the sum of classification loss and AWK loss, and sigmoid activation function $ \sigma(\cdot)$ is adopted in our experiments.

\subsection{Deep Template Matching based PAR}
\label{sec3.1}
Template matching attempts to find instances of a given template in an existing image by finding areas of high correspondence. Each layer of data in ConvNet is a three-dimensional array of size $c \times h \times w$. Inspired by Iandola et al. \cite{squeezenet}, who discard fully-connected layers and adopt fully convolutional architecture. Convolution combined with pooling is the same operation as cross-correlation in template matching. Since the output of DTM matching score only reflects the correlation between input feature maps and templates in the receptive field of convolution kernel, the output score must be a local feature. Under the premise that DTM has the correct input features and matching templates, DTM module can extract local information as needed. On the output heatmap of DTM, the score of each point corresponds to the confidence of attribute presence in a certain position. Then, we can get the predicted score of each attribute by applying pooling operation to heatmaps.

For input feature maps $F\in R^{ n\times c\times h\times w}$, most PAR works first apply Global Average Pooling (GAP) to feature maps, followed by Fully-Connected Layer (FC) for classification.
\begin{equation}
\hat{y} = \sigma(BN(W_{fc}\cdot GAP(F)))
\label{formula1}
\end{equation}
where $\cdot$ means the dot product of two matrixs, $W_{fc}$ is the weight matrix of FC layer, and $BN$ represents Batch Normalization Layer \cite{BN}. In our work, to satisfy the form of template matching, we first use templates ($1\times1$ convolution kernels) $T_c$ with the shape $(J, C, 1, 1)$ to get the heatmaps which denote the confidence of attribute presence in the corresponding position and is of size $(J,H,W)$, followed by a pooling layer for classification.
\begin{equation}
\hat{y} = \sigma(GAP(BN(T_{c}^{a}\ast F))) 
\label{formula2}
\end{equation}
where $T_{c}^{a}$ represents the templates for GAP. Without BN layer, given input feature maps $F$, we can get the same mathematical results from  Eq. \ref{formula1} and Eq. \ref{formula2}. However, BN layer can alleviate the imbalanced data problem in PAR datasets, correct the bias between data and help network convergence. Therefore, BN is widely used in the output layer of PAR. With BN layer added, we find that the Eq. \ref{formula2} method can achieve a better performance than the way in Eq. \ref{formula1}. The reason is that the number of features used in BN layer calculation are different. For Eq. \ref{formula1}, given the input $(N,C,1)$, BN layer computes statistics on $(N,1)$ slices. As for Eq. \ref{formula2}, given the input of size $(N,C,H,W)$, BN is done over the $C$ dimension, computing statistics on $(N,H,W)$ slices, which provide us more high-level features. The quantitative results will be introduced in Sec. \emph{Ablation Study} and Fig. \ref{figure4}.

Due to the reason that, some attributes require global features to predict, e.g., age, gender, shape, etc., while some attributes only need local features, e.g., hair, cloth, attachment, etc., different pooling strategies are applied to different attributes‘ heatmaps to obtain the final predicted results. Predicted heatmaps are divided into two sets: 
\begin{equation}
\hat{y} = \sigma(GAP(BN(T_{c}^{a}\ast F)) \oplus GMP(BN(T_{c}^{m}\ast F)))
\label{formula3}
\end{equation}
One is for global attributes predicted via templates $T_{c}^{a}$ and global average pooling (GAP), one is for local attributes predicted via templates $T_{c}^{m}$ and global max pooling (GMP). Finally, we concatenate ($\oplus$) two sets of results as our final predicted results.

The sigmoid weighted cross-entropy loss \cite{deepmar} is adopted as the loss of multi-class attribute classification in our work, denoted in Eq. \ref{formula4} as $L_{wce}$:  
\begin{equation}
\begin{aligned}
L_{wce} = -\frac{1}{N}\sum_{i=1}^{N}\sum_{j=1}^{J}w_j(y_{ij}\log{(\sigma(\hat{y_{ij})})} \\ 
+ (1-y_{ij})\log{(1-\sigma(\hat{y_{ij})})}), \\
\end{aligned}
\label{formula4}
\end{equation}
\begin{equation}
w_j = \begin{cases}
exp((1-p_j)/ \lambda^2), &y_{ij}=1 \\
exp(p_j/ \lambda^2), &y_{ij}=0
\end{cases}
\label{formula5}
\end{equation}
To deal with the unbalanced distribution of attributes, in Eq. \ref{formula5}, $w_j$ denotes the learning weights for $j$-th attribute, $p_j$ is the positive ratio of $j$-th attribute in training set and $\lambda$ is a tuning parameter which is set as 1 in our experiments.

\subsection{Auxiliary Supervision with AWK}
As is stated above, we use DTM to get the heatmap which represents the location of attribute presence. However, well-trained templates are hard to get in the training process. At the initial stage of training, input feature maps of DTM are features extracted from a pre-trained model without fine-tuning and templates (convolution kernels) are initialized with random parameters, in consequence, the location of maximum on the feature map is random, which will likely lead to an inaccurate localization. Further, since the backward gradients on the feature map ahead of the global max pooling layer only back-propagate gradients through a single position where forward prediction's maximum locates, some features on the positive sample image that are not related to attribute will be mistaken as positive features to update template's parameters. Also, one maximum score location can not provide enough information for template learning discriminative and generalizable features which may lead to overfitting or underfitting on the training set. To solve the above problems, we utilize an auxiliary supervision method to guide templates towards the correct location during training. We transfer the attribute recognition problem to the detection of attribute presence on heatmap with the auxiliary supervision of human pose keypoints. We use AWK to calculate loss function on heatmap to supervise templates learning. 
Since PAR datasets are not annotated with keypoints information, we need to use existing state-of-the-art human pose estimation method to generate keypoints at first. The details of generating keypoints in our work are introduced in Sec. \emph{Implementation details}.

\begin{figure}[t]
	\begin{center}
		\scalebox{1}[0.9]{
		\includegraphics[scale=0.6]{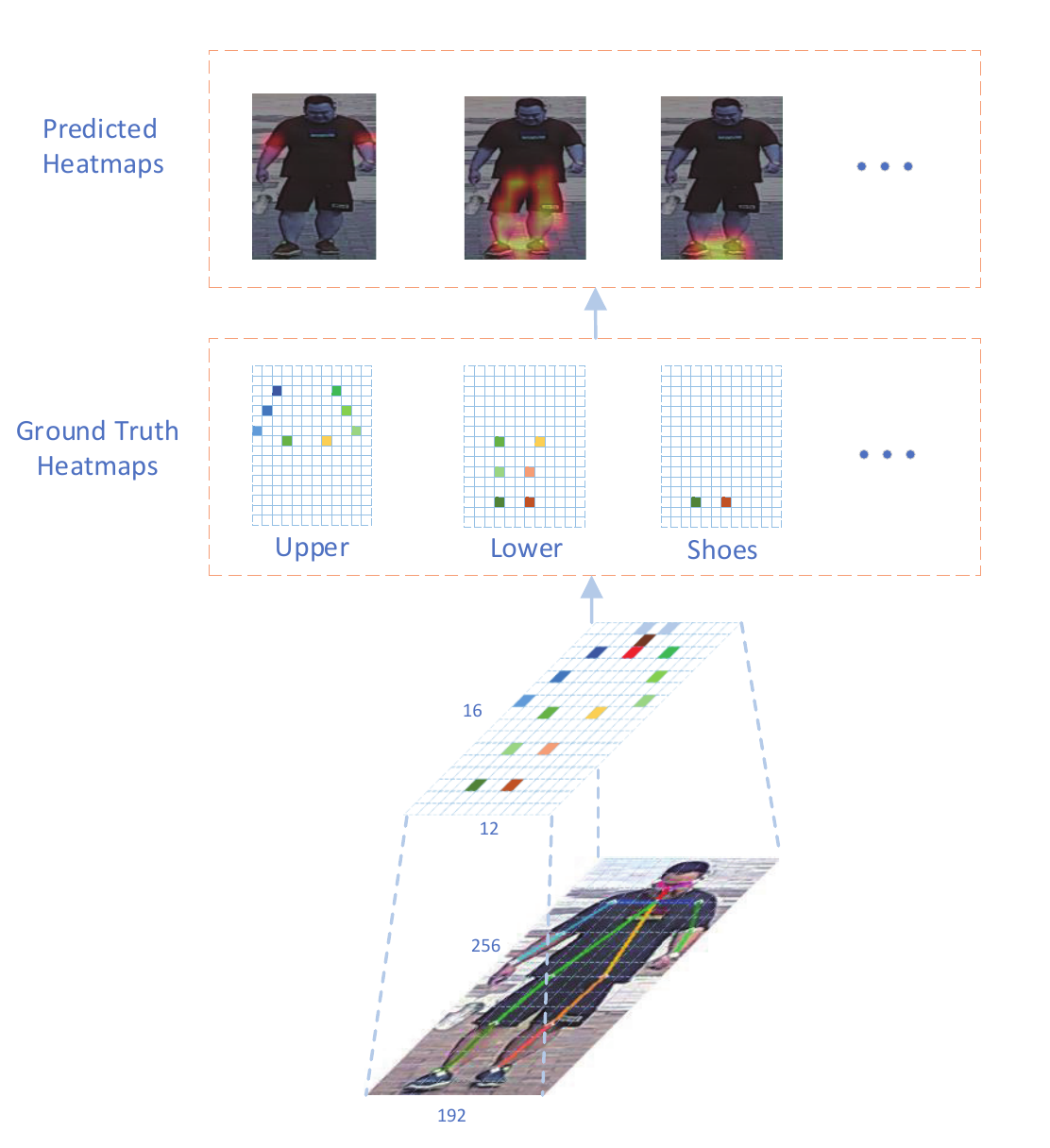}}
	\end{center}
	\caption{Details of AWK loss calculation. ``Ground truth'' heatmaps are generated based on emprical assignment of keypoints and attribute ground truth. }
	\label{fig4}
\end{figure}
The procedure of AWK supervision is illustrated in Fig. \ref{fig4}. Given input image $I \in R^{w\times h\times 3}$ and model downsample rate, we can know the predicted output heatmaps are $\hat{H}\in [0,1]^{{\frac{w}{r}}\times \frac{h}{r} \times J} $, where $r$ is the downsampling factor of model and $J$ represent number of classes. Keypoint location $(x,y)$ on the image is mapped to the location $(\lfloor \frac{x}{r}\rfloor,\lfloor \frac{y}{r} \rfloor)$ based on down stride $r$. The predicted heatmap is defined as $\hat{H}_{ij}$, and $\hat{H}_{ij}(s)$ represents the score of $\hat{H}_{ij}$ at position $s$, where $s \in \{1, 2, ..., S\}$ and $S$ is the size of heatmap, $S=\frac{w}{r}\times \frac{h}{r}$. Accroding to the keypoints assignment criteria which is human prior knowledge and set by us empirically as shown in Tab. \ref{table1}, the $j$-th attribute's AWK are defined as $P_j = \{P_{jk} |\hspace{3pt} 1 \le k \le K_j, P_{jk}\in\{1,2,..., S\}\}$, where $K_j$ is the number of keypoints we use for $j$-th attribute.

Then AWK loss is introduced to supervise templates localization of attribute-specific area: 
\begin{equation}
L_{awk} = -\frac{1}{N}\sum_{i}^{N}\sum_{j}^{J} F(\hat{H}_{ij}, y_{ij}, P_{j})
\end{equation}
\begin{equation}
F(\cdot) = \begin{cases}
\frac{1}{K_j}\sum\limits_{k=1}^{K}\log(\sigma(\hat{H}_{ij}(P_{jk})), & y_{ij}=1 \\ 
\frac{1}{S}\sum\limits_{s=1}^{S}\log(1-\sigma(\hat{H}_{ij}(s)), &  y_{ij}=0
\end{cases}
\end{equation}
\begin{table}[t]
	\begin{center}
		\scalebox{1}[0.9]{
		\begin{tabular}{c|c}
			\hline
			Attribute & Keypoints \\
			\hline
			Glasses & nose,eyes, ears \\
			Jacket & shuolders, elbows, wrists, hips\\
			Trousers & hips, knees, feet\\
			SportShoes & feet \\
			Attachments & hands \\
			Calling & hands, ears \\
			... & ... \\
			\hline
		\end{tabular}}
	\end{center}
	\caption{Examples of assignment of keypoints for attributes}
	\label{table1}
\end{table}
Our AWK loss is calculated attribute by attribute. It should be noted that when we calculate the heatmap loss of the positive sample, we only take the heatmap score at the position of keypoints into consideration, instead of calculating within the range of whole heatmap. There are two reasons for this: (1) We apply max pooling to heatmap to get the predicted score. Only one extreme position is needed rather than all locations, we do not need to care about the extracted feature from irrelevant position; (2) The down-sampled keypoints' locations are coarse-grained, at the same time, AWK supervision only plays a role in guiding templates towards interest region to extract the features of attribute, rather than detection of keypoints accurately. We have done experiments to verify that if we calculate the loss of non-keypoints on the heatmaps of positive samples as negative, it will cause our model's task in confusion and not converge. As for negative samples, we treate every positon of heatmap as negative features and calculate loss within the entire heatmaps to ensure that the accurate results can be obtained after max pooling. In the end, our AWK auxiliary supervision block is discarded at the inference stage. The overall loss function can be obtained by: 
\begin{equation}
Loss = \alpha L_{awk} + \beta L_{wce} \\
\end{equation}
where $\alpha,\beta$ are the weights for AWK and classification loss respectively.

In our experiments, we find that global-based attributes, e.g., age, gender, shape, etc., is difficult to define supervised keypoints by human, for example, the age of a person can be predicted from upper or lower body (dress style), head (face), etc. Our experiments prove that it is better for the model to learn the feature by itself than we specify regions. Therefore, we only utilize AWK supervision on local attributes in our experiments. 
\begin{table*}[htbp] 
	\centering
	\resizebox{\textwidth}{!}{
		\begin{tabular}{c|ccccc|ccccc|ccccc}
			\hline
			Dataset & \multicolumn{5}{c|}{PETA} & \multicolumn{5}{c|}{PA-100K} & \multicolumn{5}{c}{RAP}\\
			\hline
			\diagbox{Method}{Metrics} & mA & Accu & Prec & Recall & F1 & mA & Accu & Prec & Recall & F1 & mA & Accu & Prec & Recall & F1\\
			\hline\hline
			DeepMar \cite{deepmar} & 82.89 & 75.07 & 83.68 & 83.14 & 83.41 & 72.70 & 70.39 & 82.24 & 80.42 & 81.32 & 73.79 & 62.02 & 74.92 & 76.21 & 75.56\\
			Strong Baseline \cite{Rethinkingzeroshot} & 85.19 & 79.14 & \underline{87.11} & 86.18 & 86.36 & 80.50 & \textbf{78.84} & 87.24 & 87.12 & 86.78 & 80.52 & \underline{68.44} & 79.91 & 80.64 & 79.89\\  
			\hline
			HP-Net \cite{PA100K} & 81.77 & 76.13 & 84.92 & 83.24 & 84.07 & 74.21 & 72.19 & 82.97 & 82.09 & 82.53 & 76.12 & 65.39 & 77.33 & 78.79 & 78.05\\
			VeSPA \cite{vespa} & 83.45 & 77.73 & 86.18 & 84.81 & 85.49 & 76.32 & 73.00 & 84.99 & 81.49 & 83.20 & 77.70 & 67.35 & 79.51 & 79.67 & 79.59\\
			\hline
			JRL \cite{JRL} & 82.13 & - & 82.55 & 82.12 & 82.02 & - & - & - & - & - & 74.74 & - & 75.08 & 74.96 & 74.62\\ 
			GRL \cite{GRL} & \underline{86.70} & - & 84.34 & \textbf{88.82} & 86.51 & - & - & - & - & - & 81.20 & - & 77.70 & 80.90 & 79.29\\
			\hline
			PGDM \cite{PGDM} & 82.97 & 78.08 & 86.86 & 84.68 & 85.76 & 74.95 & 73.08 & 84.36 & 82.24 & 83.29 & 74.31 & 64.57 & 78.86 & 75.90 & 77.35\\
			LG-Net \cite{LGNet} & - & - & - & - &- & 76.96 & 75.55 & 86.99 & 83.17 & 85.04 & 78.68 & 68.00 & \underline{80.36} & 79.82 & 80.09 \\
			ALM \cite{alm} & 86.30 & \underline{79.52} & 85.65 & \underline{88.09} & \underline{86.85} & 80.68 & 77.08 & 84.21 & \underline{88.84} & 86.46 & \underline{81.87} & 68.17 & 74.71 & \textbf{86.48} & 80.16\\ 
			CoCNN \cite{CoCNN}& \textbf{86.97} & \textbf{79.95} & \textbf{87.58} & 87.73 & \textbf{87.65} & 80.56 & 78.30 & \textbf{89.49} & 84.36 & \underline{86.85} & 81.42 & 68.37 & \textbf{81.04} & 80.27 & \textbf{80.65}\\
			\hline\hline
			DTM (GAP) & 85.24 & 79.26 & 86.81 & 86.67 & 86.48 & \underline{80.70} & \underline{78.82} & \underline{87.37} &87.02 & \textbf{87.20} & 81.25 & \textbf{68.60} &79.91 & 81.17 & \underline{80.53}  \\
			
			DTM+AWK (GAP+GMP) & 85.79 & 78.63 & 85.65 & 87.17 & 86.11 & \textbf{81.63} & 77.57 & 84.27 & \textbf{89.02} & 86.58 & \textbf{82.04}& 67.42 & 75.87 & \underline{84.16} & 79.80\\
			\hline\\
		\end{tabular}}
		\caption{Experimental results of the proposed method and other state-of-art works on PETA, PA-100K, RAP dataset with bold \textbf{best} result and underline \underline{second best} result. Competitors are categorized into four aspects: global-based, attention-based, relation-based, and part-based.}
		\label{table2}
	\end{table*}
\section{Experiments}

\subsection{Dataset and Metrics}
We evaluate the proposed method on existing large-scale benchmark datasets, including PETA \cite{PETA}, PA-100K \cite{PA100K}, RAP \cite{RAPv1} and RAPV2\textsubscript{$zs$} \cite{Rethinkingzeroshot}. \textbf{PA100K} dataset consists of 100,000 pedestrian images from 598 scenes. The whole dataset is randomly split into training, validation, and test sets with a ratio of 8:1:1. Each image has 26 commonly used attributes. \textbf{PETA} dataset consists of 19000 images labeled with 61 binary attributes. Same with previous work \cite{PETA}, we use the 35 selected attributes for evaluation, and partitions provided by work \cite{deepmar}.  \textbf{RAP} has in total 41585 images, each of which is annotated with 72 attributes as well as viewpoints, occlusions, body parts information. We adopt the same partition setting as the previous work \cite{deepmar}. \textbf{RAPv2$_{zs}$} is newly proposed by Jia et al. \cite{Rethinkingzeroshot} who argue that images of the same pedestrian identity in training set and test set are extremely similar, leading to overestimated performance of state-of-the-art methods. To solve this problem in existing PAR dataset, based on the pedestrian identity label provided in RAPv2 \cite{RAPv2} dataset, they propose RAPv2$_{zs}$ dataset, which has no overlap pedestrian identity between the train set and test set. Zero-shot setting based dataset is more realistic and practical in real scenario. RAPv2$_{zs}$ has 26632 images in total and 54 selected attributes, our experiments follow the division as proposed.

Two types of metrics proposed by Li et al. \cite{RAPv1} are adopted for fair evaluation, i.e. label-based metric and instance-based metrics. (1) Label-based metric, i.e. mean accuracy (mA), computes the mean accuracy of positive and negative examples for each attribute. And then calculate the average of all attributes. (2) Instance-based metrics, i.e. accuracy, recall, precision, and F1 score. Since there is a balance between precision and recall, F1 might be a better measurement.

\subsection{Implementation Details}
In our experiments, we adopt ResNet50 \cite{resnet} pre-trained from ImageNet as backbone network. The original down stride of ResNet50 is 32, we reduce it to 16 by changing the stride of the last bottleneck to 1. This change balances the requirement of high-level features for classification and fine-grained feature for localization of attribute presence. We use Zhou el al. \cite{CenterNet} pre-trained model which uses hourglass as backbone \cite{hourglass} and is trained on COCO dataset \cite{coco} to generate 17 pedestrian keypoints on PAR dataset images. We assign these keypoints to different attribute empirically as preliminary of training, examples are illustrated in Tab. \ref{table1}. SGD optimizer is employed for training with the momentum of 0.9, and weight decay equals 0.0005. The initial learning rate is set to 0.01 and batch size equals 64. We adopt the image with a size of 256 $\times$ 192 as input and apply random horizontal mirroring and random crop as data augmentation. The weights of both loss functions are set to 1. Our experiment is implemented with Pytorch and trained end-to-end. 

\begin{table}
	\centering
	\resizebox{8cm}{!}{
		\begin{tabular}{c|ccccc}
			\hline
			Dataset &                    \multicolumn{5}{c}{RAPv2 Zero-shot} \\
			\hline
			\diagbox{Method}{Metrics} & mA & Accu &  Prec  &Recall & F1\\ 
			\hline
		     MsVAA \cite{MsVAA} *& 71.32 & 63.59 &   77.22 & 76.62 & 76.44\\
           VAC \cite{VAC} *  & 70.20 & \textbf{65.45} &   79.87 & 76.65  & 77.07\\
           ALM \cite{alm} *  &  71.97 & 64.52  &   77.28 &  77.74 & 77.06 \\
		Strong Baseline \cite{Rethinkingzeroshot} &    70.83 & 63.63 & \textbf{82.28} & 72.22 &   76.34   \\ \hline
		DTM (GAP) & 72.94 & 64.52 & 77.19 & 77.94 & 77.56\\
		DTM+AWK(GAP + GMP) & \textbf{73.60} &  65.13  &     73.26  & \textbf{84.05} & \textbf{78.29} \\ \hline
		\end{tabular}}
		\caption{Experimental results on RAPv2\textsubscript{$zs$}. $Results^*$ of MsVAA, VAC, and ALM are reimplemented by Jia et al. \cite{Rethinkingzeroshot}, since the dataset is newly proposed. Our experiments adopt the same setting as their works for fair comparison. Best results are in \textbf{bold}.}
		\label{table3}
	\end{table}
\subsection{Experimental Results}
We compare our proposed method with a number of state-of-the-art works, e.g. DeepMar \cite{deepmar}, HP-Net \cite{PA100K}, JRL \cite{JRL}, VeSPA \cite{vespa}, GRL \cite{GRL}, PGDM \cite{PGDM}, LG-Net \cite{LGNet}, CoCNN \cite{CoCNN}, ALM \cite{alm} and Strong Baseline \cite{Rethinkingzeroshot}. All experiments are evaluated under standard protocol \cite{deepmar}. The results of ours DTM, DTM with AWK, and serveral state-of-the-art methods are listed in Tab. \ref{table2}.

\textbf{RAP:} On RAP dataset, DTM with AWK surpasses all previous methods in label-based metric with 82.04\% in mA. Our DTM achieves the best accuracy and a competitive F1 score which is slightly lower than the best CoCNN.

\textbf{PA-100K:} On PA-100K dataset, DTM with AWK outperforms all previous work on label-based metric (mA) and recall with 81.63\% and 89.02\%, respectively. Our DTM achieves the best in F1 score, and second-best in accuracy and precision.

\textbf{PETA:} As shown in Tab. \ref{table2}, we find that DTM with AWK does not achieve the best performance than existing state-of-the-art methods, e.g. CoCNN, ALM, and GRL in label-based metric mA. However, we still get comparable results. We think the main reason may be the low-resolution images in PETA datasets. Due to the low-resolution, some wrong human pose estimation may lead the templates to locate in the wrong place. 

\textbf{RAPv2\textsubscript{$zs$}:} Since it is a newly proposed dataset, it is unlikely to compare with previous pedestrian attribute methods, we refer to Jia et al.'s \cite{Rethinkingzeroshot} reimplemented results as our comparison benchmark. As shown in Tab. \ref{table3}, DTM with AWK outperforms other works, achieves best on both label-based metric and instance-based metric. The improvement of performance on reasonable RAPv2\textsubscript{$zs$} datasets proves that our proposed method is more effective and has stronger generalization ability in realistic scenario. 

The promising experimental performance on four datasets shows the superiority of our methods. For DTM, we have achieved significant improvement and competitive performance. For DTM with AWK, our method achieves state-of-the-art performance on label-based metric, benefit from AWK modules. While maintaining a high recall rate, our precision rate is also acceptable. High recall rate and mA verify our initial thought that the auxiliary supervision helps the template learning hard samples.

\subsection{Complexity Analysis}
\begin{table}[h]
	\centering
	\resizebox{8cm}{!}{
		\begin{tabular}{l|c|cc|c}
			\hline  
			Method & Backbone & Params(M) & GFLOPs (G) & mA \\
			\hline
			DeepMar \cite{deepmar} & CaffeNet & 58.5 & 0.72 & 73.79 \\
			PGDM \cite{PGDM} & CaffeNet & 87.2 & $\approx$ 1 & 74.31\\
			\hline
			GRL \cite{GRL} & Inception-V3 & \textgreater50 & \textgreater10 & 81.20\\
			VeSPA \cite{vespa} & BN-Inception & 17.0 & \textgreater3 & 77.70\\
			LGNet \cite{LGNet} & BN-Inception & \textgreater 20 & \textgreater 4 & 78.68\\
			ALM \cite{alm} & BN-Inception & 17.1 & 1.95 & 81.87 \\
			DTM + AWK (GAP + GMP) & BN-Inception & 10.9 & 1.43 & 81.30 \\
			\hline
			Strong Baseline \cite{Rethinkingzeroshot} & ResNet50 & 23.7 & 2.69 & 80.52\\
			DTM (GAP) & ResNet50 &  23.7 & 2.69 & 81.25\\
			DTM + AWK (GAP + GMP) & ResNet50 & 23.7 & 4.09 & 82.04 \\
			\hline
		\end{tabular}}
		\caption{Complexity Analysis on RAP dataset.}
		\label{ComplexityAnalysis}
	\end{table}

The complicated design of model may bring improvements on performance, but what cannot be ignored is that when the model is deployed on devices with limited resources, computing resources may not meet the needs of real-time inference. The DTM and AWK modules we proposed only increases the training time slightly and do not bring in additional computation and memory cost at inference stage. As shown in Tab. \ref{ComplexityAnalysis}, compared with the methods which use BN-Inception as backbone, our methods have lower Params and FLOPs. Compared with strong baseline which uses ResNet50 as backbone network, our GFLOPs is higher with AWK module, because we have modified the down stride to 1 of the last bottleneck in our experiments and increase the computation of convolution operation, as illustrated in Sec. \emph{Implementation details}. Since both hardware and software have optimized the operation of convolution layer, the increase in GFLOPs caused by convolution does not lead to a proportional increase in inference time. Without AWK module, our DTM does not bring any extra complexity. In general, DTM and AWK can be integrated on most of the commonly used backbones to achieve performance improvement without extra computation cost.
\subsection{Ablation Study}

To better demonstrate the effectiveness and advantage of DTM and AWK auxiliary supervision training block, we apply component-wise ablation studies to explicitly address the contribution of each block. The ablation study on RAP \cite{RAPv1} dataset shows that our DTM and AWK auxiliary training methods bring about a promising improvement.

\begin{figure}[h]
	\begin{center}
		\scalebox{0.78}[1]{
		\includegraphics[scale=0.20]{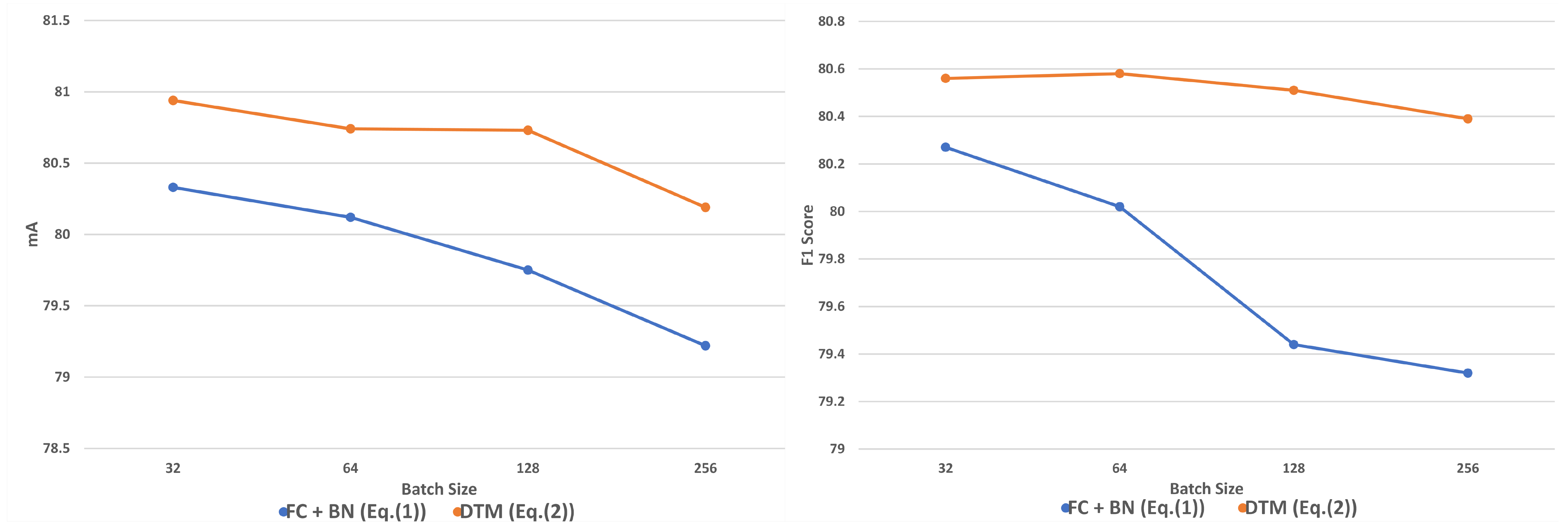}}
	\end{center}		
	\caption{Performance comparison between DTM and conventional `classifier' on different batch size settings.}
	\label{figure4}
\end{figure}

Most PAR methods adopt fully-connected layer as `classifier'. We replace conventional operation with DTM and make performance improved apparently. We argue that the number of features used in statistics of batch normalization influence the performance in Sec. \emph{Proposed Method}. In general, batch normalization layer can correct the bias between output data. When we have more number of features involved in calculation, the network is easier to converge. Increasing the training batch size can also increase the number of features used in BN layer statistics. However, increasing batch size has its pros and cons. We conduct experiments under different batch size settings to verify our thoughts and measure the impact of batch size. As shown in Fig. \ref{figure4}, as the batch size increases, the performance of two methods decrease to varying degrees, however, DTM is better than the conventional method in any cases on both label-based metric (mA) and instance-based metric (F1 score).   

\begin{figure}[htb]
	\begin{center}
		\scalebox{0.95}[1]{
		\includegraphics[scale=0.20]{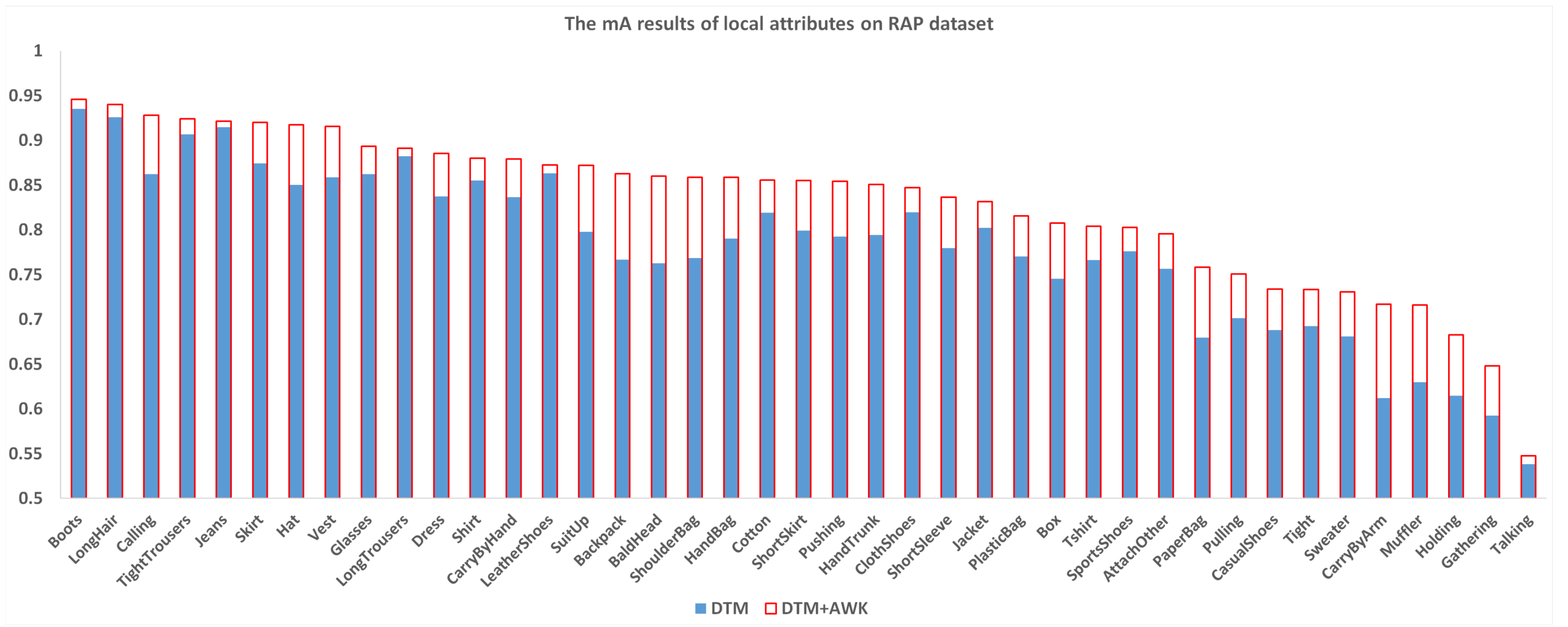}}
	\end{center}		
	\caption{Attribute-wise mA comparison between DTM with and without auxiliary training on RAP dataset.}
	\label{figure2}
\end{figure}

In our experiment settings, we use AWK auxiliary supervision for local attributes. Fig. \ref{figure2} shows the attribute-wise mA comparison between DTM with and without the AWK supervision block. As is shown, DTM with auxiliary training block achieves significant improvement on most attributes, 4\%-10\% improvement has been made on each attribute. 

Further, we provide quantitative evaluation to investigate the performance of DTM and AWK in Tab. \ref{abstudy}. Replace conventional `classifier' with DTM (GAP) leads to performance improved apparently, but replacement with DTM (GMP) leads to a minor drop. According to our observation, attributes based on global features drop a lot, e.g., age, gender, and shape, which means local features are insufficient to estimate global attributes. Therefore, mixed pooling strategy is adopted to get the best performance. When both DTM and AWK modules are employed, the label-based metric (mA) can be further improved. The results of ablation study show that DTM can make an improvement overall metrics, and AWK can improve label-based metric effectively.
\begin{table}[h]
	\centering
	\resizebox{8cm}{!}{
		\begin{tabular}{l|ccccc}
			\hline
			Method & mA & Accu & Prec & Recall & F1\\
			\hline
			FC + BN & 80.05 & 66.79 & 78.58 & 79.82 & 79.19\\
			DTM (GMP) & 77.72 & 65.37 & 77.49 & 78.99 & 78.24 \\
			DTM (GAP) & 81.25 & 68.60 & 79.91 & 81.17 & 80.53 \\
			DTM (GAP+GMP) & 81.33 & 68.51 & 79.53 & 81.42 & 80.47\\
			DTM+AWK (GAP+GMP) & 82.04 & 67.42 & 75.87 & 84.16 & 79.80 \\
			\hline
		\end{tabular}}
		\caption{Ablation study on RAP dataset.}
		\label{abstudy}
	\end{table}
\subsection{Further Anaysis and Discussions}
Compared with other state-of-the-art approaches, which bring in complex modules, our method only brings in minor computational inefficiency while training. The auxiliary supervision training block is easier to generalize on commonly used backbones, e.g., ResNet, Inception, DLA, etc. Although training with auxiliary keypoints makes a significant localization and performance improvement, empirically assignment of keypoints to attributes still have drawbacks. The assignment criteria are from human prior knowledge, not learned by model, hence, many special circumstances may not be considered. For example, we assign hands to attachment attributes, but in some special cases, attachments may be far from body or occluded by background. Moreover, the performance of attribute recognition model is largely affected by the accuracy of human pose estimation method. Fortunately, the current human pose model can produce good enough results which help us improve the performance of attribute recognition. For further study, we suggest other information may be used as auxiliary supervision tools including but not limited to keypoints, edge, and pixel-wise segmentation. 

\section{Conclusion}
In this paper, we introduce two modules to improve pedestrian attribute recognition performance, namely DTM and AWK. We first discuss that local attributes are related to local regions and propose DTM module to extract local features. To further improve the performance, we proposed AWK auxiliary supervision method which uses human pose keypoints as prior knowledge to guiding the templates toward learning local cues. With DTM and AWK modules, our method outperforms a wide range of existing pedestrian attribute recognition methods on large-scale benchmark datasets. Moreover, we have provided feature maps visualization which demonstrates our method's effectiveness intuitively. Compared with previous state-of-the-art methods, our method does not introduce additional computational complexity which makes it easier to be applied in realistic applications.

\bibliography{mybib}
\bibliographystyle{aaai21}
\end{document}